\documentclass[a4paper, 9pt, twoside]{IEEEtran}

\usepackage[utf8]{inputenc}
\usepackage[T1]{fontenc}

\usepackage{csquotes}
\usepackage[
    maxbibnames=6,
    minbibnames=1,
    backend=biber,
    url=true,
    eprint=true,
    style=ieee,
    dashed=False, 
]{biblatex}
\usepackage{booktabs}
\usepackage{longtable}
\usepackage{graphics}
\usepackage[binary-units=true]{siunitx}
\usepackage{tabularx}
\usepackage{multirow}
\usepackage{amsmath}
\usepackage{mathtools}
\usepackage{xfrac}
\usepackage{dsfont}
\usepackage{svg}
\usepackage{todonotes}

\usepackage{graphicx}
\usepackage[acronym]{glossaries}
\usepackage[export]{adjustbox}

\usepackage{hyperref}
\usepackage[capitalise]{cleveref}
\usepackage[shortcuts]{extdash} 

\newacronym{gan}{GAN}{generative adversarial network}
\newacronym{dem}{DEM}{digital elevation model}
\newacronym{radar}{radar}{radio detection and ranging}
\newacronym{lidar}{lidar}{light detection and ranging}
\newacronym{cnn}{CNN}{convolutional neural network}
\newacronym{fid}{FID}{Fréchet inception distance}
\newacronym{spade}{SPADE}{SPatially-Adaptive (DE)ormalization}
\newacronym{sims}{SIMS}{semi-parametric image synthesis method}
\newacronym{vae}{VAE}{variational autoencoder}
\newacronym{miou}{mIoU}{mean intersection over union}
\newacronym{iou}{IoU}{intersection over union}
\newacronym{sar}{SAR}{synthetic aperture radar}
\newacronym{gpu}{GPU}{graphics processing unit}
\newacronym{eec}{EEC}{Enhanced Ellipsoid Corrected}
\newacronym{in}{IN}{instance normalization}
\newacronym{bn}{BN}{batch normalization}
\newacronym{relu}{ReLU}{rectified linear unit}
\newacronym{rcs}{RCS}{radar cross section}
\newacronym{rmse}{RMSE}{root-mean-square error}
\newacronym{ssim}{SSIM}{structural similarity index measure}
\newacronym{igbp}{IGBP}{international geosphere-biosphere programme}

\usepackage[english]{babel}

\newcommand{\MAT}[1]{\mathbf{\expandafter\MakeUppercase\expandafter{#1}}}

\title{Synthesizing Optical and SAR imagery from\\ Land Cover Maps and Auxiliary Raster Data}

\author{Gerald Baier, Antonin Deschemps, Michael Schmitt, and Naoto Yokoya
    \thanks{G. Baier is with the RIKEN Center for Advanced Intelligence Project, Tokyo 103--0027, Japan (e-mail: \mbox{gerald.baier@riken.jp}).}
    \thanks{A. Deschemps is with the SERPICO Team of Inria, 35042 Bretagne-Atlantique, France (e-mail \mbox{antonin.deschemps@inria.fr}).}
    \thanks{M. Schmitt is with the Department of Geoinformatics, Munich University of Applied Sciences, 80335 Munich, Germany (e-mail \mbox{michael.schmitt@hm.edu}).}
    \thanks{N. Yokoya is with the Department of Complexity Science and Engineering, Graduate School of Frontier Sciences, the University of Tokyo, Chiba 277--8561, Japan, and also with the RIKEN Center for Advanced Intelligence Project, Tokyo 103--0027, Japan \mbox{(e-mail: yokoya@k.u-tokyo.ac.jp)}}
}
\date{}

\begin{document}

\maketitle

\begin{abstract}

We synthesize both optical RGB and \acrshort{sar} remote sensing images from land cover maps and auxiliary raster data using \acrlongpl{gan}.
In remote sensing many types of data, such as \acrlongpl{dem} or precipitation maps, are often not reflected in land cover maps but still influence image content or structure.
Including such data in the synthesis process increases the quality of the generated images and exerts more control on their characteristics.

Spatially adaptive normalization layers fuse both inputs, and are applied to a full-blown generator architecture consisting of encoder and decoder, to take full advantage of the information content in the auxiliary raster data.

Our method successfully synthesizes medium (\SI{10}{\meter}) and high (\SI{1}{\meter}) resolution images, when trained with the corresponding dataset.
We show the advantage of data fusion of land cover maps and auxiliary information using \acrlong{miou}, pixel accuracy and \acrlong{fid} using pre-trained U-Net segmentation models.
Handpicked images exemplify how fusing information avoids ambiguities in the synthesized images.
By slightly editing the input our method can be used to synthesize realistic changes, i.e., raising the water levels.

The source code is available at \url{https://github.com/gbaier/rs_img_synth} and we published the newly created high-resolution dataset at \url{https://ieee-dataport.org/open-access/geonrw}.

\end{abstract}


\section{Introduction}

\glsreset{gan}
\glsreset{sar}
Remote sensing enables researchers and scientists to detect changes, monitor areas or measure physical properties.
By analyzing the acquired images they can deduce what happened and is happening on the ground.
In this manuscript we wish to reverse this process.
Given some abstract information or constraining physical parameters, can we synthesize a remote sensing image as it would be acquired by the sensor?
Our motivation is not purely philosophical.
Often these parameters can be more easily altered than directly modifying the actual remote sensing image.
For example, it is straightforward to change input constraints by raising the sea level or swapping the climate of Moscow and Cairo, giving us a glimpse of how things may look like in the future or at the very least serve as an interesting thought experiment.
Doing such editing in the image domain is much harder.
Synthesizing multiple images, each under different constraints, permits us to visualize changes, which in reverse can be used for analyzing and training change detection algorithms.

Recent advances in image synthesis were largely driven by \Acrfullpl{gan}~\cite{schmidhuber1992learning,NIPS2014_goodfellow_gan}, which exploit both the generative and discriminative power of neural networks.
\acrshortpl{gan} pit two neural networks against each other: the generator network tries to fool the discriminator network by creating fake data that is indistinguishable from real data.
Both generator and discriminator are trained concurrently, gradually improving each other.
This idea becomes even more intriguing if one exerts some form of control over the generator and discriminator by conditioning both on a common state or variable~\cite{mirza2014conditional}.
With the discriminator knowing what input the generator received and what the corresponding real data looks like, it can now guide the generator into transforming the input into something which resembles the real data distribution more closely and adheres to the common information.
\acrshortpl{gan} were notoriously difficult to train, often resulting in mode-collapse, where the discriminator starts to memorize all real images and no longer provides any useful guidance to the generator, in which case the training stops.
There have been recent advances to avoid this behavior, either by handicapping the discriminator's training~\cite{miyato_spectral_normalization_gan_2018,Liu_2019_ICCV} or augmenting the dataset~\cite{zhao2020differentiable,karras2020training} to prevent its memorization.

A common application of \acrshortpl{gan} is synthesizing new or altering existing images and has seen tremendous progress in various computer vision applications.
They can transfer the styles of paintings to photographs~\cite{gatys_image_style_transfer_2016}, colorize images~\cite{zhang2016colorful}, translate between domains~\cite{isola_pix2pix_2016}, such as creating images from sketches, or synthesize completely new images.
Examples of the last category are creating close-to photorealistic portraits~\cite{Karras_2019_CVPR,karras2020analyzing} or turning class labels into images that show the corresponding content~\cite{miyato_spectral_normalization_gan_2018,zhang_self_attention_gan_2019,brock_biggan_2019}.
Taking image synthesis from class labels one step further are methods that condition on segmentation maps~\cite{wang_pix2pix_hd_2018,park_spade_2019,liu_cond_conv_img_syn_2019_NEURIPS}, which additionally exert spatial control, that is where in the image to put what kind of content.
Together with image translation approaches these methods are the most relevant for our work.

\begin{figure*}
    \centering
    \includegraphics[width=0.9\textwidth]{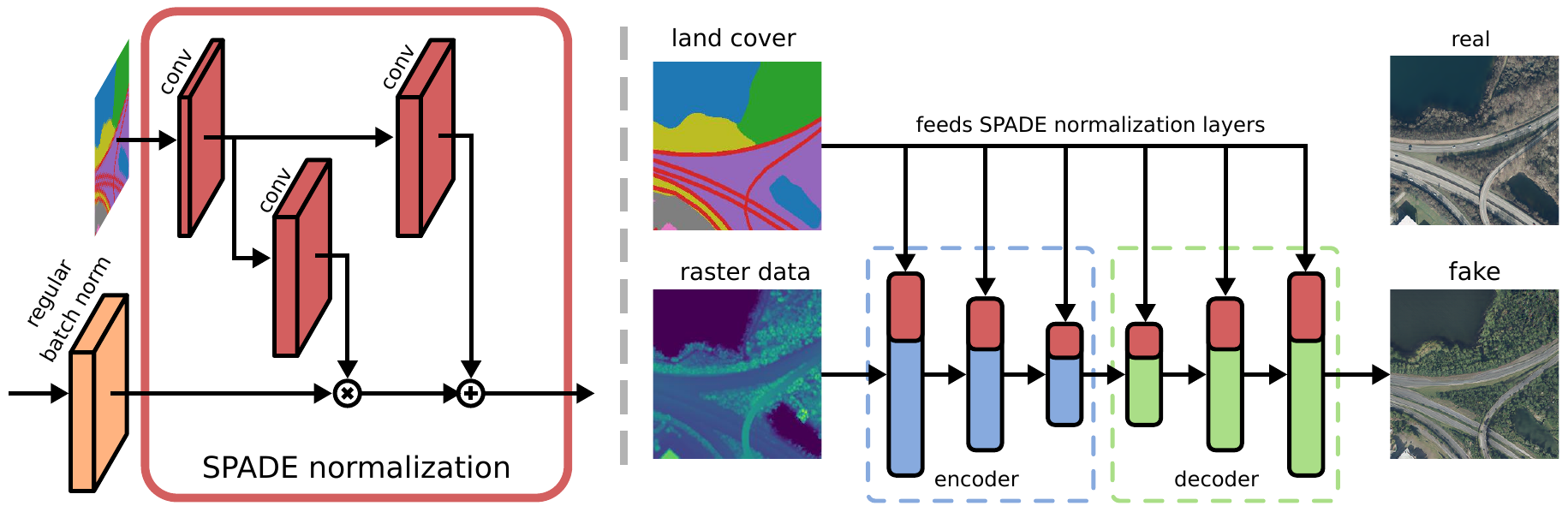}
    \caption{Generator architecture (right side) for synthesizing images from raster data and land cover maps.
    Semantic information is fed into the generator through SPADE~\cite{park_spade_2019} normalization layers (left side).
    In our experiments this architecture performed better than simply concatenating both inputs and using a conventional generator, or a more involved scheme that fuses two generators, each dedicated to one type of input.
    By replacing all SPADE normalization layers with regular batch normalization, the generator turns into a conventional generator~\cite{isola_pix2pix_2016} that only uses raster data as input.
    Conversely, removing the encoder simplifies the generator to the SPADE~\cite{park_spade_2019} generator for segmentation maps.}%
    \label{fig_gen_arch}
\end{figure*}

A couple of works already concern themselves with image synthesis and translation within the scope of remote sensing imagery.
\Cite{abady2020ganmulti} covers two topics: synthesizing Sentinel-2 multispectral images, although not conditioned on a segmentation map; and style transfer between vegetation and bare lands, also for Sentinel-2 images.
\Cite{andrade2020gan_historical} essentially translates historical maps to overhead RGB images by merging the created images of multiple generators, each trained with a focus on different types of landscapes.
In the related field of domain adaption,\ \cite{tasar2020colormapgan} proposes a simple generator architecture to rid satellite images of color distortions, as a result of atmospheric and other environmental effects.
Colorizing \acrshort{sar} images or translating them into artificial optical images to make them more easy interpretable by laymen is proposed by~\cite{schmitt2018colorizing_sar} and~\cite{he2018multi,Fuentes_Reyes_2019}, respectively.

Although most of the research works employ \acrshortpl{gan} for image synthesis, \acrlong{vae}s~\cite{kingma2014autoencoding,Kingma_2019} are another approach, that reach state-of-the-art performance~\cite{razavi2019generating}.

With this paper, we want to bring image synthesis to the field of remote sensing.
Taking image synthesis methods in computer vision as a starting point, the most relevant use segmentation maps as input, which exert control on the output on a pixel-level.
Yet, in remote sensing there is often auxiliary data available that can make this task easier.
These can be a large variety of maps, images or physical properties, including \glspl{dem}, precipitation maps, local heat maps, and the concentration of gases.
Such quantities are often not reflected in land cover maps but still influence image content or structure.
We thus propose a generator architecture that merges the abstract information contained in land cover maps and additional data sources, which we assume can be encoded in a raster matching the land cover map.
In this paper we restrict ourselves to synthesizing RGB and \gls{sar} images from land cover maps and one more data source.
In one of our datasets this is a \acrshort{dem}, which helps with synthesizing vegetation and buildings.
For the other datasets, where we do not have a \acrshort{dem} of comparable resolution, we fuse \acrshort{sar} with the land cover map, to imprint the structure of the \acrshort{sar} image (e.g., buildings, roads) on the synthesized RGB outputs.
Our method should still be general enough to translate between other image domains as well.

Even such a limited scope enables some interesting applications.
The earlier mentioned look into the future is still possible by light editing of the input, e.g., the transformation of forest into farm land.
In case sensor artifacts, noise or other perturbations get synthesized, such images might help to design new restoration algorithms.
In a similar vein, image synthesis could also help with the actual training of new neural networks by generating training data.
As an example, numerical simulations of rain fall can produce flood and debris flow maps, which can then in turn be used to train machine learning models~\cite{yokoya2020breaking}.
Such a method can be extended with our work by synthesizing images from said maps.
These can then be used in reverse for end-to-end training of new models.

The paper's major contributions are as follows
\begin{enumerate}
    \item a \acrshort{gan}-based image synthesis method that merges semantic information and raster data to generate RGB or \acrshort{sar} images,
    \item the publication of a high-resolution dataset for image synthesis and segmentation~\cite{geonrw_dataset}, and
    \item analyzing challenges and pitfalls when synthesizing remote sensing imagery.
\end{enumerate}

Our PyTorch implementation will be available at \mbox{\url{https://github.com/gbaier/rs_img_synth}} and the dataset is already published at \mbox{\url{https://ieee-dataport.org/open-access/geonrw}}.

\section{Method}

\subsection{Fusion of semantic and depth information}

In contrast to image translation~\cite{isola_pix2pix_2016} or semantic image synthesis~\cite{wang_pix2pix_hd_2018,park_spade_2019,liu_cond_conv_img_syn_2019_NEURIPS} algorithms, our objective requires a generator architecture that consumes two very different types of inputs
\begin{enumerate}
    \item abstract, high-level information as land cover maps, and
    \item unprocessed, auxiliary raster data.
\end{enumerate} 

We tried concatenating both inputs and feeding them to a conventional generator~\cite{isola_pix2pix_2016}, but this process was prone to generate artifacts when synthesizing buildings.
Presumably, since the normalization of the semantic maps causes issues~\cite{park_spade_2019}.
Furthermore, the generator's capacity is partially wasted on encoding and passing along the already abstract, high-level semantic information.
We also experimented with fusing essentially two separate generators, one for the raster data and the other for semantic information, in a separate upsampling path at various layers.
Again this approach turned out to be less performant than expected.
Most likely due to the fact, that the two generators and upsampling path require additional \acrshort{gpu} memory, necessitating a reduction of their capacity, which in turn hampers performance~\cite{brock_biggan_2019}.
We arrived at the conclusion, that semantic information should only undergo little processing before being introduced to the generator, and that a simple, basic architecture but with high capacity is superior to more elaborate schemes with fewer filters per layer.

In lieu of these two options, we thus opted to extend \acrshort{spade}~\cite{park_spade_2019} to a complete full-blown generator.
\acrshort{spade} synthesizes images from semantic maps alone.
It infuses the information contained in semantic maps by special normalization layers, depicted on the left of \cref{fig_gen_arch}, which replace all normalization layers in the generator.
Since segmentation maps already encode distilled, high-level information, the encoder in typical image translation generators~\cite{wang_pix2pix_hd_2018} is superfluous and can be removed, leaving only the decoder paired with \acrshort{spade} normalization layers.
In our use-case, an encoder is still needed to process the information contained in the raster data.
We thus employ a conventional generator architecture, consisting both of an encoder and a decoder, where the encoder only receives the raster data as input, and just like in \acrshort{spade}, semantic maps pass trough a separate path directly into the normalization layers (\cref{fig_gen_arch}, right side).
The following section describes the generator architecture in detail.

\subsection{Generator architecture}

We base the generator on~\cite{johnson_style_transfer_2016}, which is also the underlying architecture of the generators in~\cite{isola_pix2pix_2016} and~\cite{wang_pix2pix_hd_2018}.
Yet, we use a slightly modified decoder, so that the upsampling path matches the recent \acrshort{spade} architecture.
That is, we interleave ResNet-Blocks~\cite{he_identity_mappings_2016,he_resnet_2016} with nearest-neighbor upsampling to avoid checker-board artifacts~\cite{odena2016deconvolution}.
As mentioned in the previous section, we introduce semantic information from land cover maps into the generator by replacing all its normalization layers with \acrshort{spade} layers~\cite{park_spade_2019}.
In total, there are four down- and upsampling stages, which proved to be sufficient in our experiments for synthesizing images of size $256 \times 256$ or $512 \times 512$.

\Cref{fig_gen_arch} shows a sketch of the proposed generator architecture.
It has the typical encoder-decoder structure known from~\cite{isola_pix2pix_2016}, which can be used to synthesize images from raster data, but infuses information from discrete land cover maps through its \acrshort{spade} normalization layers.
This generator architecture has convenient properties.
Replacing all \acrshort{spade} normalization layers with conventional batch normalization results in a generator that synthesizes images from raster data alone.
Conversely, removing the encoder simplifies the generator to the SPADE~\cite{park_spade_2019} generator with only segmentation maps as input.
These properties allow us to easily investigate the benefit of synthesizing images from different or multiple sources.

\Cref{tab_generator} lists all layers in detail.
In the encoder strided convolutions $C$ downsample the feature maps, followed by either batch~\cite{Szegedy_2016} or \acrshort{spade}~\cite{park_spade_2019} normalization layers and \acrfullpl{relu}.
Identical to~\cite{johnson_style_transfer_2016}, there is a body of cascaded ResNet blocks to further encode the image without any downsampling.
As mentioned above, nearest neighbor upsampling in the decoder return the feature maps to it's original dimension.
The final layer consists again of a three-by-three convolution and a hyperbolic tangent activation function.
The final number of output channels $\xi$ depends on the image type that is to be synthesized, i.e., three for RGB, one for single-pol \acrshort{sar} and two for dual-pol.

\begin{table}[thb]
\centering
\newcommand{\ra}[1]{\renewcommand{\arraystretch}{#1}}
\newcommand{\rotentry}[2]{\parbox[t]{2mm}{\multirow{#1}{*}{\rotatebox[origin=c]{90}{#2}}}}
\ra{1.2}
\caption{Generator architecture.
    Strided convolutions $C^{\downarrow 2}$ downsample feature maps, which is reversed by nearest neighbor upsampling in the decoder.
The number of output channels in the final layer depends on the datatype that is synthesized.}
    \begin{tabular*}{\columnwidth}{l @{\extracolsep{\fill}} l r}
    \toprule
                      & Layers                                      & \# out channels \\
    \midrule
\rotentry{5}{Encoder} & $C_{7 \times 7}$, ReLU                      &   64 \\
                      & $C_{3 \times 3}^{\downarrow 2}$, batch or \acrshort{spade} normalization, ReLU &  128 \\
                      & $C_{3 \times 3}^{\downarrow 2}$, batch or \acrshort{spade} normalization, ReLU &  256 \\
                      & $C_{3 \times 3}^{\downarrow 2}$, batch or \acrshort{spade} normalization, ReLU &  512 \\
                      & $C_{3 \times 3}^{\downarrow 2}$, batch or \acrshort{spade} normalization, ReLU & 1024 \\
    \midrule
                      & $9 \times$ ResNet block                     & 1024 \\
    \midrule
\rotentry{5}{Decoder} & Nearest Neighbor $\uparrow 2$, ResNet block & 512 \\
                      & Nearest Neighbor $\uparrow 2$, ResNet block & 256 \\
                      & Nearest Neighbor $\uparrow 2$, ResNet block & 128 \\
                      & Nearest Neighbor $\uparrow 2$, ResNet block &  64 \\
                      & $C_{3 \times 3}$, Tanh                      & $\xi$ \\
    \bottomrule%
\end{tabular*}%
\label{tab_generator}
\end{table}

\subsection{Discriminator architecture}

The discriminator is identical to the multiscale discriminator proposed in~\cite{wang_pix2pix_hd_2018}.
As in~\cite{park_spade_2019}, we found two scales for images with resolutions of $256 \times 256$ and three for $512 \times 512$ to be perfectly sufficient for obtaining satisfying results.
\Cref{tab:discriminator} lists all layers for a single scale, \gls{in} follows all but the first convolutional layers.
The final output is the average of the discriminators' outputs at multiple scales.

\begin{table}[tb]
\centering
\newcommand{\ra}[1]{\renewcommand{\arraystretch}{#1}}
\ra{1.2}
\setlength{\tabcolsep}{10pt}
\caption{Discriminator architecture (identical to SPADE~\cite{park_spade_2019}).
\acrlong{in} (\acrshort{in}) is used in all but the first layer.
}
\begin{tabular}{l r}
    \toprule
    Layer & \# out channels \\
    \midrule
    $C_{4 \times 4}^{\downarrow 2}$, Leaky ReLU & 64 \\
    $C_{4 \times 4}^{\downarrow 2}$, \acrshort{in}, Leaky ReLU & 128 \\
    $C_{4 \times 4}^{\downarrow 2}$, \acrshort{in}, Leaky ReLU & 256 \\
    $C_{4 \times 4}$, \acrshort{in}, Leaky ReLU & 512 \\
    $C_{4 \times 4}$ & 1 \\
    \bottomrule%
\end{tabular}%
\label{tab:discriminator}
\end{table}

\subsection{Losses and training}

We directly adopt the loss terms from \acrshort{spade}~\cite{park_spade_2019}.
Let $x$ denote the generator input, i.e., the \acrshort{dem} and land cover map, and $y$ the desired real output, i.e., \acrshort{sar} or RGB images.
$G$ and $D$ represent the generator and discriminator, and $D_k$ the discriminator's $k$-th feature layer.
The generator loss consists of the regular \acrshort{gan} loss and a discriminator feature matching loss~\cite{improved_gan_training_salimans_nips_2016} between real and synthesized images
\begin{align*}
    \mathcal{L}_G =  \underset{x,y \sim q(x,y)}{-\mathrm{E}} \left\{ D(G(x), x) + \sum_k \lVert D_k(G(x), x) - D_k(y, x) \rVert_1 \right\},
\end{align*}
with $q$ denoting the data distribution.
For brevity we disregarded the multiple scales of the discriminator.
The discriminator itself is optimized with the Hinge loss
\begin{align*} 
    \mathcal{L}_D =  \underset{x,y \sim q(x,y)}{\mathrm{E}} \{ &\min\left(0, -1 + D(y, x) \right)  \\ 
      + & \min\left(0, -1 - D(G(x), x) \right)\}.
\end{align*}

\subsection{Peculiarities of remote sensing data}

Remote sensing data, be it multispectral or \acrshort{sar}, are actual measurements of physical properties from carefully calibrated instruments.
This sets them apart from cellphone or camera photographs often used in computer vision.
As a result the dynamic range of remote sensing data is much greater than regular photographs.
\acrshortpl{dem} range from a couple of meters below sea level, the Netherlands spring to mind, to the height of the Mount Everest with \SI{8848}{\meter}.
Regarding \acrshort{sar}, \gls{rcs} or the backscatter coefficient $\sigma_0$ can fluctuate between \SI{-30}{\decibel} to \SI{10}{\decibel} by just moving a couple of meters from buildings to roads.
In the case of Sentinel-2 multi-spectral imagery, L1C data measures top-of-atmosphere reflectances, which are represented by integer values ranging from 0 to 10,000 (for reflectances from 0\% to 100\%, respectively). For Sentinel-2 L2A data, the pixels refer to bottom-of-atmosphere reflectance.
Although in our experiments the dynamic range of the various data types did not cause any problems when using them as input, the story is different for the output.
Due to the architecture of the generator, with a final $\tanh$-layer (\cref{tab:discriminator}) the output is limited to the interval $(-1, 1)$ and we have to properly normalize all data types to fall in this range.
This include clipping them to a sensible range, and for \acrshort{sar} taking the logarithm to make the data more amenable for synthesis.
This is similar to the conversion from linear to decibel when plotting any \acrshort{sar} amplitude or intensity data.
The experimental section details the exact normalization for each dataset.

\section{Experiments}

With our experiments we set out to demonstrate
\begin{enumerate}
    \item the synthesis of convincing RGB and \acrshort{sar} imagery,
    \item the benefit of fusing land cover maps and auxiliary raster data,
    \item and that editing the input results in sensible changes.
\end{enumerate}

\subsection{Datasets}

We employ two datasets of high and medium resolution.
Both contain optical and \gls{sar} imagery to highlight the proposed method's versatility.

\paragraph{GeoNRW} consists of orthorectified aerial photographs, \acrshort{lidar} derived \acrshortpl{dem}, land cover maps with 10 classes and TerraSAR-X spotlight acquisitions over the German state North Rhine-Westphalia~\cite{geonrw_dataset}.
Since urban areas are the most challenging to synthesize we focused on gathering data from urban centers such as the Rhein-Ruhr area, Düsseldorf or Cologne.
The TerraSAR-X images were acquired from German Aerospace Center (DLR) by means of a research proposal, and can thus unfortunately not be made publicly available.
All other data are freely available as part of the open data program of North Rhine-Westphalia~\cite{opengeodata_nrw}.
We however refine the coarse land cover maps of the open data program with building footprints and roads from OpenStreeMap~\cite{OpenStreetMap}.
Other preprocessing consisted of resampling the \SI{10}{\centi\meter} resolution aerial photographs to \SI{1}{\meter}, taking the first \acrshort{lidar} return while averaging within \SI{1}{\square\meter} to arrive at the same resolution as the photographs, and rasterizing the land cover maps.
We directly download geocoded and terrain corrected TerraSAR-X spotlight \gls{eec} acquisitions and resample them to the same grid.

\begin{figure}
    \centering
    \includegraphics[width=\columnwidth]{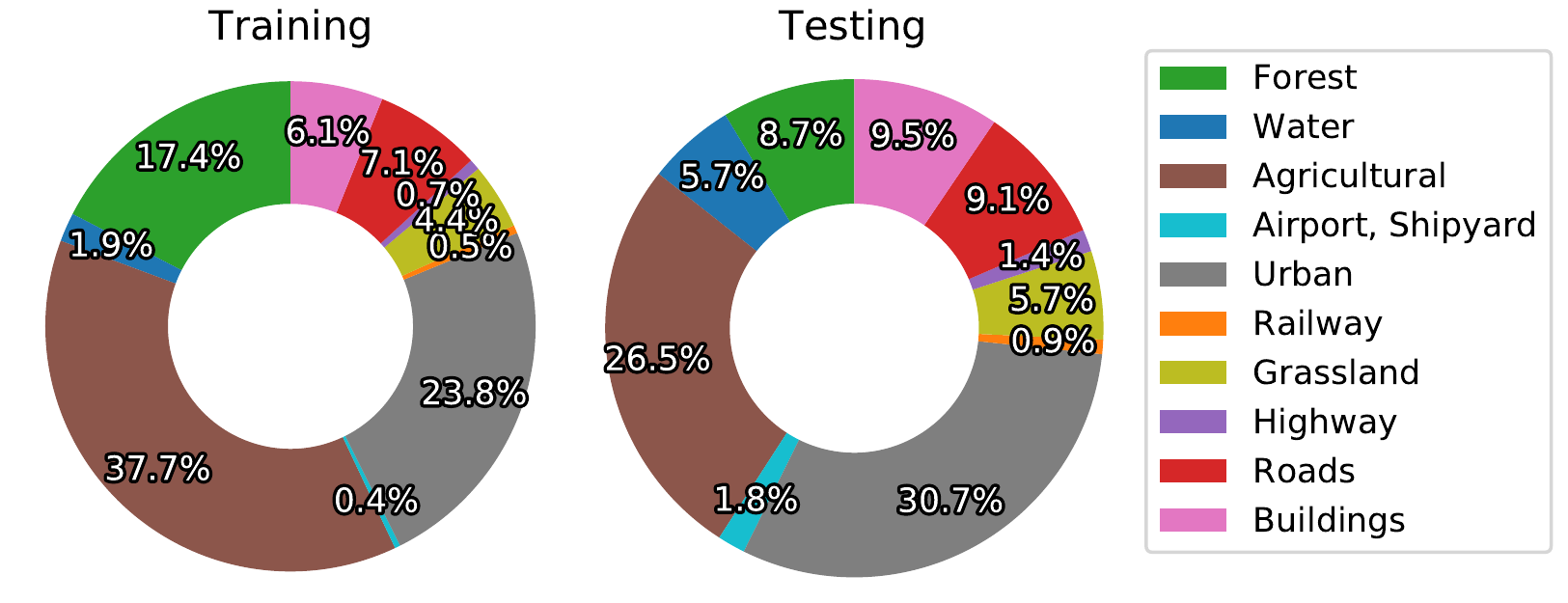}
    \caption{Class statistics of the GeoNRW dataset.
        There is severe class-imbalance, with man-made structures being in the minority.
        This can affect both, segmentation accuracy and synthesis quality.
    }%
    \label{fig_nrw_stats}
\end{figure}

We end up with 7782 tiles of aerial photographs, land cover maps and \acrshortpl{dem} of size 1000 $\times$ 1000, of which 485 make up the test set and the rest the training set.
\Cref{fig_nrw_stats} shows the composition of the training and testing datasets.
Classes are severely imbalanced with man-made structures being less common than natural surfaces.
Since the TerraSAR-X archive does not contain data for all these tiles the \acrshort{sar} dataset is smaller and only consists of 2980 tiles, 281 in the test set and the other 2699 in the training set.
This also slightly alters the class statistics.

Normalizing the TerraSAR acquisitions consisted of the conversion to decibel, subsequently dividing by a factor of $100.0$ resulted in a range of values that could be clipped to the interval $(0, 1)$ and serve as a target for the generator.
The other data types did not require normalization.
\acrshortpl{dem} are encoded as the height in meters above sea-level, with no extreme outliers and being used as input only, the networks learn a proper normalization themselves.
The aerial photographs are comparable to images in computer vision in terms of dynamic range and do not require any particular processing.

\paragraph{IEEE GRSS Data Fusion Contest 2020 (DFC2020)}

The DFC2020 aimed at the training of models for automated large-scale land cover mapping from coarse, noisy labels.
For this purpose, the simplified land cover scheme of the \acrfull{igbp}, consisting of generic 10 classes, was adopted.
We use the high-resolution validation/testing labels of the data fusion contest~\cite{yokoya2020dfc}, together with the corresponding Sentinel-1 dual-pol and Sentinel-2 acquisitions where only 8 of these 10 classes are present.
The dataset contains images with \SI{10}{\meter} ground sampling distance from a diverse set of locations and scene types around the world.
The official split of the data fusion contest assigns 5128 tiles to the testing set and 986 to the validation set.
We follow the same assignment for our training and testing sets, respectively.
Just like the GeoNRW dataset, the DFC2020 is imbalanced~\cite{schmitt_2020_dfc2020}.

Although the DFC2020 dataset does not include a \acrshort{dem} of comparable resolution, it can still provide some additional insights, when compared to the GeoNRW dataset.
It features a greater variety of locations and scene types and it is interesting to see whether these can be successfully captured by \acrshortpl{gan}.
Similar considerations are valid for its dual-pol Sentinel-1 data, in contrast to the single-pol TerraSAR-X data in the GeoNRW dataset.
In addition to synthesizing RGB and \acrshort{sar} from land cover maps alone, we also fuse \acrshort{sar} and the land cover maps to generate RGB images using our approach.

We clip the Sentinel-1 \acrshort{sar} data to the interval $(-20, 5)$, subsequently normalizing them so that all values fall into the range $(0, 1)$.
Similarly, when extracting the RGB bands from the Sentinel-2 bands, we limit them to the interval $(0, 3500)$, which results in sensible images when plotting them, and normalize them to the same interval as the Sentinel-1 data.

\subsection{Evaluation schemes and metrics}

Judging the quality of synthesized images is an intricate problem and still an active topic of research.
Besides visual inspection we perform a quantitative analysis using U-Net~\cite{ronneberger2015unet} segmentation networks, trained on the corresponding dataset.
We exclusively use real data for pre-training, which consequently biases the resulting networks in favor of this data distribution and makes them sensitive to shifts from it.

Thus, comparing the segmentation accuracy for real and synthesized images can serve as a metric for the domain gap between both image distributions.
The argument is that more realistic looking synthesized images perform better, since the pre-trained model only knows real data and is sensitive to changes.

We additionally use a slight modification of the \gls{fid}~\cite{Heusel_fid_NIPS2017} for both datasets to evaluate the quality of synthesized images.
\Gls{fid} is a refinement of the inception score~\cite{improved_gan_training_salimans_nips_2016} and compares mean and covariance of an Inception-v3~\cite{Szegedy_2016} network's (pre-trained on ImageNet~\cite{imagenet_cvpr09}) intermediate features for real and synthesized images.
The domain gap between ImageNet and remote sensing images prohibits to directly compute \acrshort{fid}.
Instead, analogously to~\cite{bau2019seeing}, we extract the intermediate features of the corresponding pre-trained U-Nets and compute the Fréchet distance~\cite{frechet1957distance} between real and synthesized images.

Classic figures of merit such as \acrfull{rmse} or \acrfull{ssim} are not applicable for image synthesis, since they focus on low-level local statistics and not on high-level semantic information.
As an example in computer vision, cars or clothes can have a variety of color, which invariably result in bad \acrshort{rmse} or \acrshort{ssim} scores if their color does not match the original, even though the generated image might look convincingly realistic.

\subsection{Implementation details and training regimen}

We apply spectral normalization~\cite{miyato_spectral_normalization_gan_2018} to all convolutional layers in the discriminator and generator.
Adam~\cite{adam_kingma_2015}, parameterized with $\beta_1 = 0$ and $\beta_2 = 0.9$ trains generator and discriminator with individual learning rates~\cite{Heusel_fid_NIPS2017} ($0.0001$ and $0.0004$) with the Hinge-loss~\cite{lim_2017_geometric_gan,miyato_spectral_normalization_gan_2018} for 200 epochs.
\acrshort{gan}-training is done on 8 NVIDIA V100 GPUs with \SI{16}{\giga\byte} VRAM each and a batch size of 32.
The GPUs share batch statistics using synchronized batch normalization.
We train the U-Net segmentation networks using Adam, with $\beta_1=0.9$ and $\beta_2=0.999$, a learning rate of $0.0002$, and a batch size of 32 for 100 epochs using cross entropy loss on a single NVIDIA V100 GPU with \SI{16}{\giga\byte} VRAM.
All networks are implemented in PyTorch~\cite{NEURIPS2019_9015}.

\begin{figure*}
    \centering
    \includegraphics[width=\textwidth]{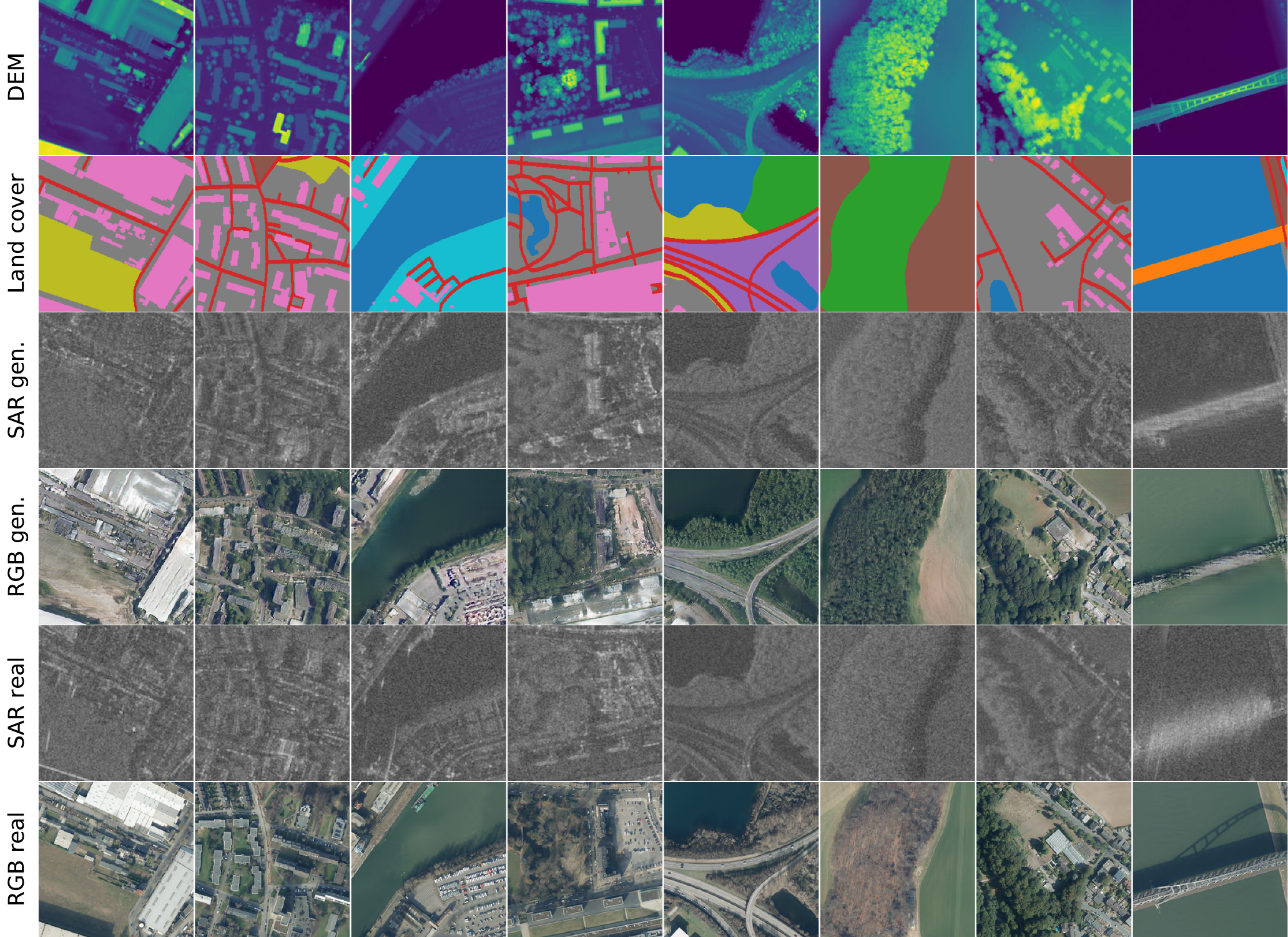}
    \caption{Synthesizing high-resolution optical and TerraSAR-X spotlight images from \acrshortpl{dem} and land cover maps.
        The third and fourth row show synthesized examples with their corresponding real images in the fifth and sixth row.
        The synthesized images look convincing.
        In particular, the generator learns to add realistic shadows to both optical and \acrshort{sar} images, which are also consistent for the whole image.
        However, man-made structures with their complex but regular shapes make it easier to distinguish between real and synthesized images upon close inspection.
        The last column shows an example where our approach fails to synthesize a convincing image.
        Bridges, like other man-made structures are difficult to synthesize, and there are not many examples in the dataset, leading to comparably poor results.
}%
    \label{fig_nrw_comp}
\end{figure*}

\subsection{Synthesizing remote sensing imagery}

\Cref{fig_nrw_comp} shows synthesized RGB and \acrshort{sar} images for the GeoNRW dataset, with \acrshort{dem}s and land cover maps as the input.
There is good correspondence between real and synthesized images.
Remarkably, the generator also synthesizes realistic shadows, both for optical and \acrshort{sar} images, that are consistent for the entire image and match building heights.
Also of note are seasonal changes.
As the aerial images were acquired between spring and autumn, trees change their leafs, the colors of which can range from green to brownish.
The same variety is captured by the synthesized images, which do not necessarily correspond to the season of their real counterpart, but still look realistic.
When closely zooming in, real and synthesized images can still be distinguished.
Artificial structures like buildings, bridges or roads are more easily identified as fake than natural land covers, such as forest, grassland or water.
We hypothesize that in addition to getting the texture right, artificial structures require a certain geometric regularity, which human observers can more easily identify if it is amiss.
Another problem facing these kind of objects is that they are not well represented in the dataset.
The last column serves as an example, where the generator fails to synthesize a realistic looking bridge.

\begin{figure*}
    \centering
    \includegraphics[width=\textwidth]{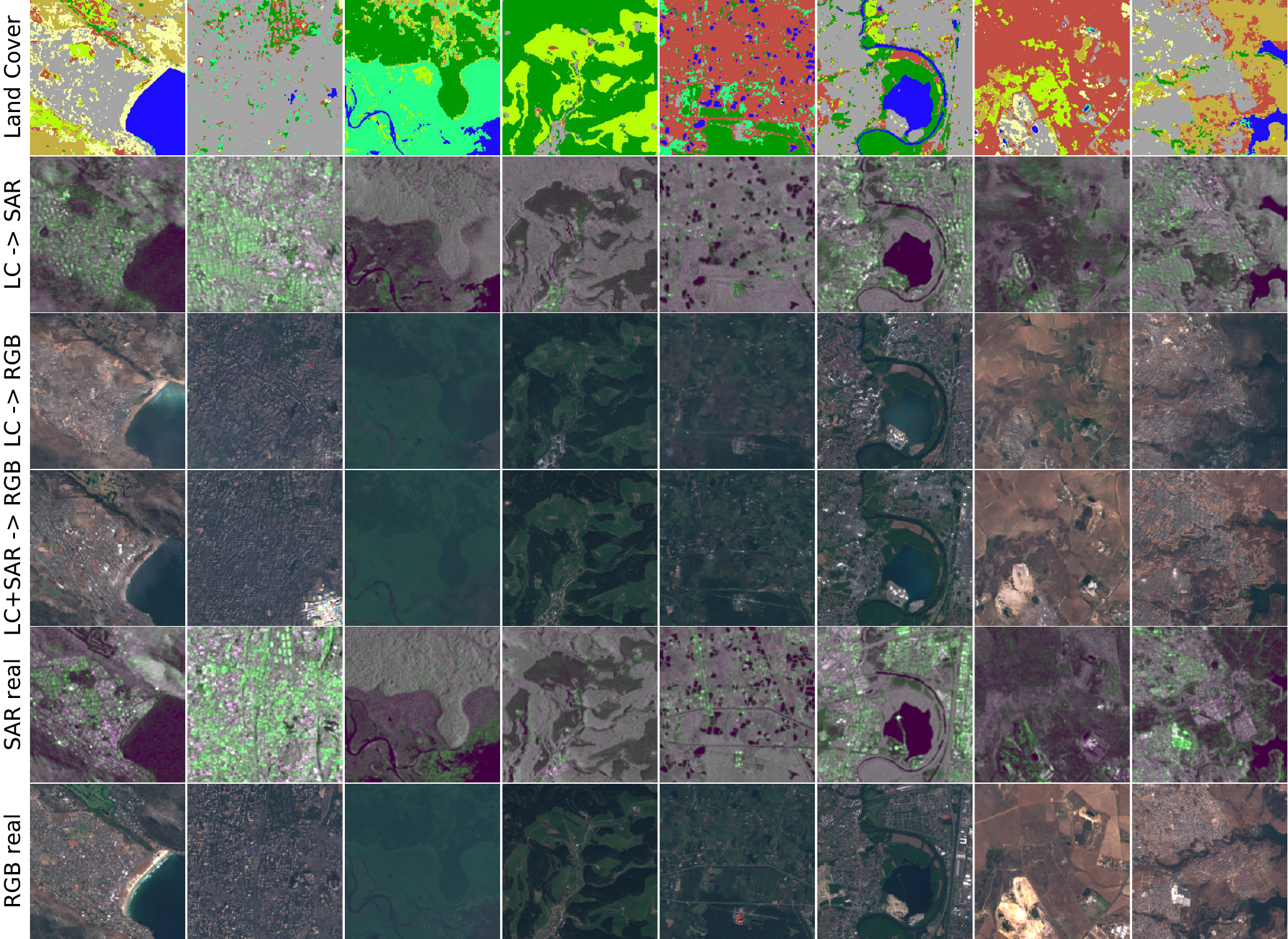}
    \caption{Synthesizing medium-resolution Sentinel-1 dual-pol and Sentinel-2 (only RGB) images from land cover maps alone and with \acrshort{sar} as an additional input.
        We use the dataset from the 2020 IEEE Geoscience and Remote Sensing Society data fusion contest~\cite{rha7-m332-19}.
        The second, third and fourth row show synthesized examples with good agreement to their corresponding real images in the fifth and sixth row.
        Urban areas and agricultural fields are an exception, where it is difficult for the network to infer their layout from land cover maps alone.
        In this case, having \acrshort{sar} as an additional input helps to transfer the \acrshort{sar} image's structure and provides some guidance.
    }%
    \label{fig_dfc_sen12}
\end{figure*}

The DFC2020 dataset exhibits more diversity in terms of classes and locations.
We employ our method to synthesize RGB, created from the corresponding Sentinel-2 bands, and dual-pol \acrshort{sar} images.
As inputs we use land cover maps alone, but also fuse them with \acrshort{sar} using the proposed architecture to give an impression on how auxiliary data can help when synthesizing images.
\Cref{fig_dfc_sen12} shows that there is good correspondence between real and synthesized images.
One caveat is the generation of urban areas from land cover maps alone.
In contrast to the GeoNRW dataset, the land cover maps of the DFC2020 dataset do not contain any structural information, which assigns the task of placing buildings or roads entirely to the generator.
Compounded by the fact that street layouts and building types and placement differ wildly between cities, as they depend on a city's history, land marks such as hills and rivers, but also building codes etc., it is especially hard to come up with a realistic layout at a large scale.
Using \acrshort{sar} as an auxiliary input imprints its structure on the synthesized images, resulting in street layouts that more closely match the original images.

\begin{figure}
    \centering
    \includegraphics[width=\columnwidth]{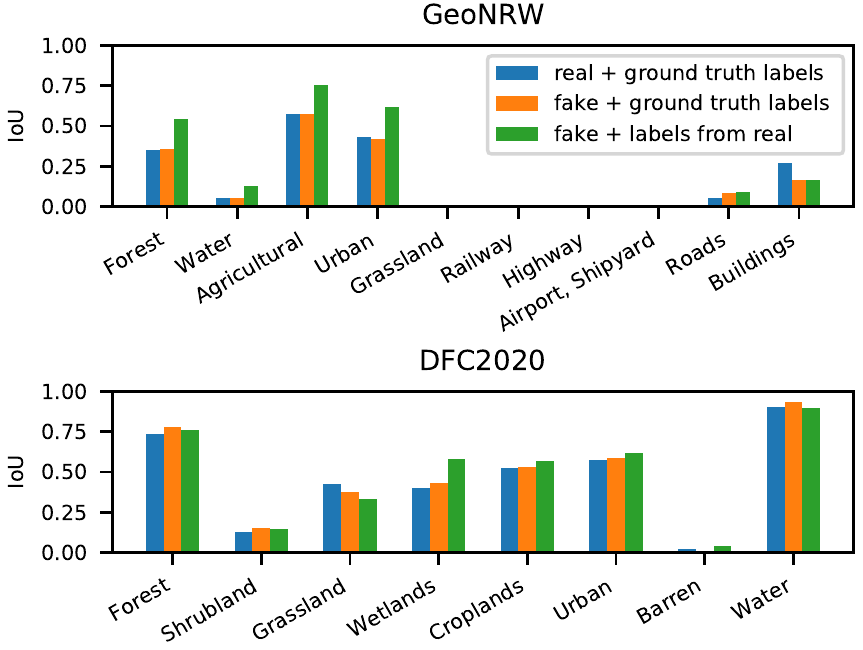}
    \caption{Intersection-over-Union for the segmentation maps produced by U-Net, pre-trained on real RGB images.
        The segmentation network is either fed real or synthesized images.
        We compare the resulting land-cover maps against the ground truth labels.
        In addition, to account for label-noise, we directly compare the segmentation maps obtained from real and fake images.
        Segmentation accuracy is strongly correlated with class-imbalance (see \cref{fig_nrw_stats} and~\cite{schmitt_2020_dfc2020}).
    }%
    \label{fig_ious}
\end{figure}

As a quantitative evaluation we compare the segmentation results obtained both from real and synthesized images in terms of \gls{iou}.
To better gauge the impact of label noise we also use the segmentation maps obtained from real data as a reference ground truth when analysing the generated images.
\Cref{fig_ious} shows the obtained \acrshortpl{iou} for the individual classes, when synthesizing RGB images from land cover maps alone (DFC2020) and in conjunction with \acrshortpl{dem} (GeoNRW).
For both datasets the segmentation accuracy is severely affected by class imbalance in the training dataset.
More sophisticated training procedures, e.g., using weighted cross entropy or focal loss~\cite{lin_2017_focal_loss} could alleviate this problem, but are outside the scope of this paper.

With respect to the ground truth labels, classification is comparable for real and synthesized images, indicating that the quality of synthesized images is good enough not to confuse a network pre-trained on real data, which conversely should make it feasible to use synthesized images for training and real data during inference~\cite{yokoya2020breaking}.
One notable exception are buildings in the GeoNRW dataset.
This confirms our observation, that buildings are the most difficult class to synthesize for the generator.
Presumably, due to their large variance in texture and shape.

\begin{figure}[tb!]
    \centering
    \includegraphics[width=\columnwidth]{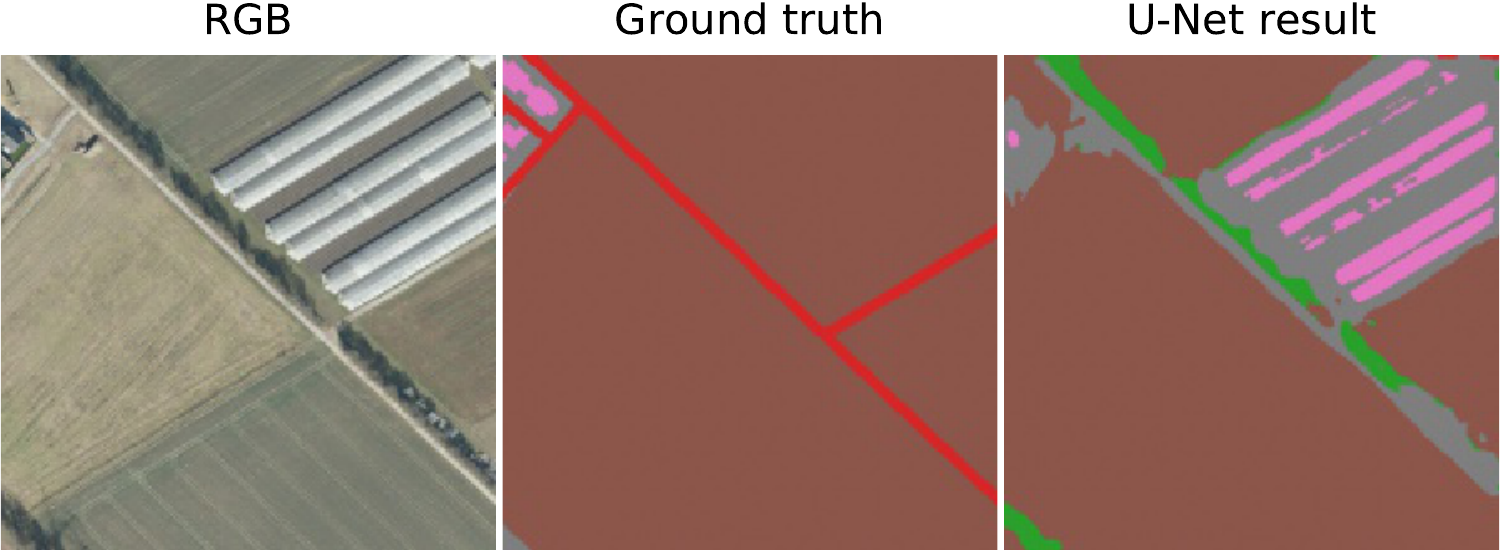}
    \caption{Example of label noise in the GeoNRW dataset.
        The coarse ground truth does not include the greenhouses' building footprints and the trees lining the road, both are picked up by the segmentation network.
    }%
    \label{fig_nrw_label_noise}
\end{figure}

Using the segmentation results obtained from real data as a reference for the synthesized images leads to higher \gls{iou} for the GeoNRW dataset.
This suggests a certain degree of label noise, which fits the relatively coarse labeling for certain land covers.
\Cref{fig_nrw_label_noise} shows an example, where the ground truth labeling does not fit the corresponding RGB image.
Trees lining the road and greenhouses are not represented in the ground truth segmentation map.

\subsection{Fusion of input sources}

This section shows the benefit of fusing input sources over just relying on one of them alone.
As mentioned in the method section, the proposed generator architecture can easily be simplified to rely on segmentation maps or unprocessed data arrays alone during synthesis.
In the former case the simplified generator is equivalent to \acrshort{spade}~\cite{park_spade_2019}, in the latter to Pix2Pix~\cite{isola_pix2pix_2016}.

\begin{figure*}
    \centering
    \includegraphics[width=\textwidth]{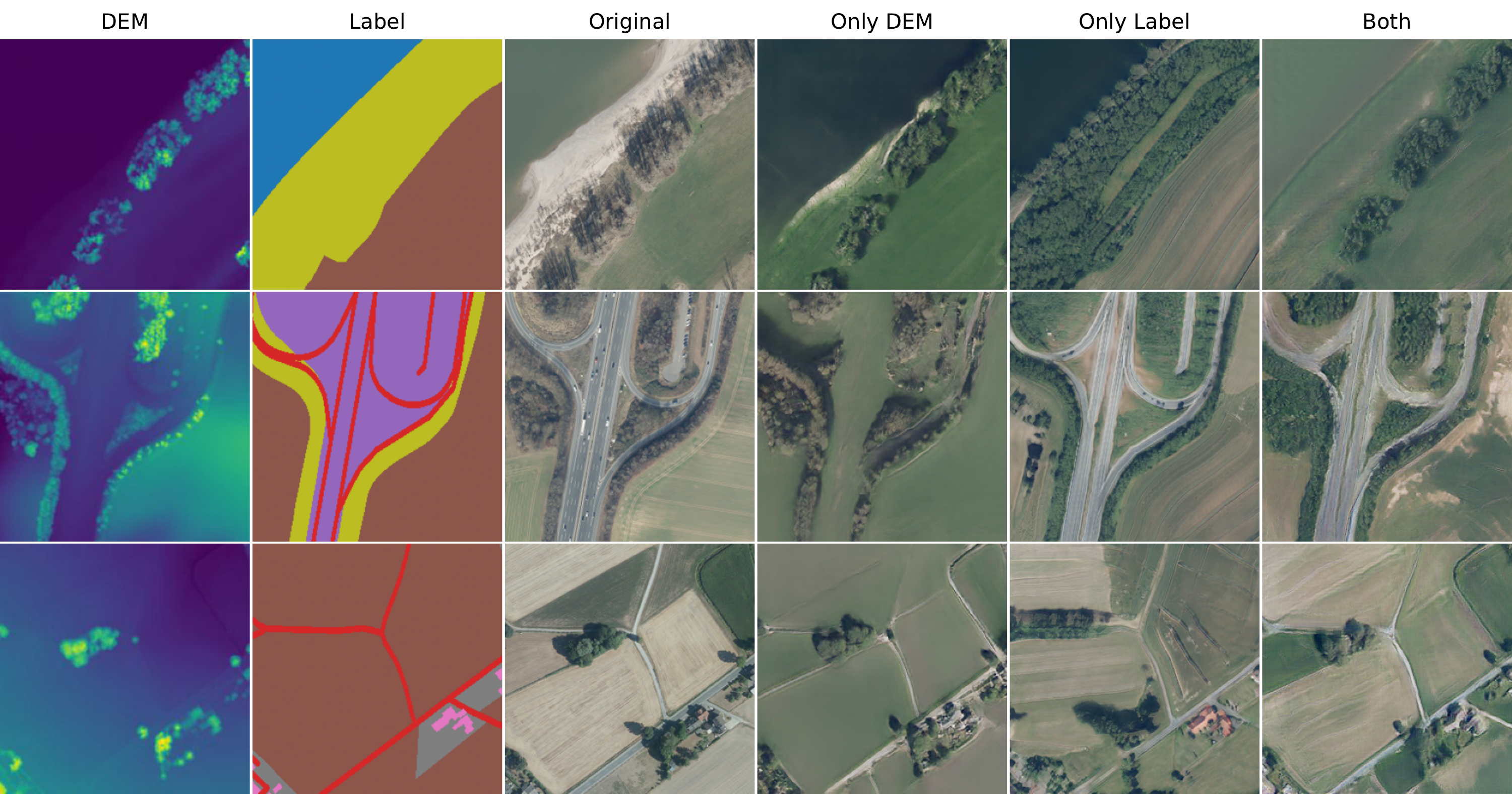}
    \caption{Using both land cover maps and \acrshortpl{dem} as input for the generator avoids ambiguities in the output, such as confusing flat regions like water, roads or agricultural fields.
    Furthermore, since the land cover maps of the GeoNRW dataset are not very detailed, the generator learns to imagine and add small details when it only receives land cover maps as input, such as the trees in the last row.
    This does not reflect the actual ground truth images, even though they show up at realistic locations, i.e., lining roads.
}%
    \label{fig_nrw_diff_sources}
\end{figure*}

We show that synthesizing images jointly from land cover maps and \acrshortpl{dem} from the GeoNRW dataset helps to avoid ambiguities in the generated image.
Three handpicked examples in \cref{fig_nrw_diff_sources} demonstrate how land cover maps and \acrshortpl{dem} complement each other.
From \acrshortpl{dem} alone, without additionally providing land cover maps, the generator can not distinguish flat regions, such as water, roads or agricultural fields.
This leads to results where these types of classes are swapped and mixed up.
Conversely, the land cover maps are not particularly detailed, i.e., they lack individual trees, or class boundaries are not well defined (for example between grassland and forest).
This results in the generator to learn to imagine small details or randomly choosing what to synthesize.
The last row in \cref{fig_nrw_diff_sources} serves as an interesting example, where the generator places trees next to roads, which certainly is an assumption based on reality, but does not reflect the actual ground truth image.
Moreover, the \acrshort{dem} occasionally delineates boundaries between objects, even if these are present in the land cover map.
Note that such boundaries, essentially edge maps, can help the generation quality by offering additional guidance was already observed in~\cite{tang2020edge}.

\Cref{tab_miou_pix_acc_fid} lists the \acrfull{miou}, pixel accuracies and \acrshortpl{fid} of our quantitive analysis, obtained using pre-trained U\=/Net models.
In addition to the GeoNRW dataset, where, as before, we synthesize RGB images from land cover maps and \acrshortpl{dem}, or one of them alone, we also use the DFC2020 dataset as a test case.
For lack of a \acrshort{dem} of comparable resolution, we resort to using \acrshort{sar} as an auxiliary input.
We also compare our method with two other generator architectures
\begin{enumerate}
    \item simply concatenating both inputs and using the Pix2Pix generator, and
    \item using the Pix2Pix generator for the \acrshort{dem}, the SPADE generator for the land cover map and fusing their decoders at all layers by concatenation in an essentially third generator. This architecture's larger memory footprint required a reduction by 25\% of the generator's capacity.
\end{enumerate}
Again, to account for label noise, the table includes results with respect to the ground truth labels and with respect to the segmentation maps obtained from real data.
For \acrshort{fid}, we of course compare the activations of the real and synthesized images.

Generally speaking, regarding \acrshort{miou} and pixel accuracies, fusion is advantageous when comparing to the segmentation obtained from real data, whereas synthesizing from labels alone performs better when comparing to the ground truth.
Again, this disparity can be explained by label-noise.
Although fusion of land cover maps with auxiliary information provides additional guidance, these two sources of information will actually contradict each other if mistakes were made during labeling.
However, when synthesizing from labels alone, and coming full circle when creating labels using a segmentation network, there is no distracting auxiliary input.

Regarding \acrshort{fid}, the output is less clear.
Contrary to \acrshort{fid} in computer vision, applying this metric in remote sensing faces two issues
\begin{enumerate}
    \item There is no dataset that is even remotely comparable in size to ImageNet~\cite{imagenet_cvpr09}.
        Learned representations will thus be less refined than in computer vision.
    \item Inception-V3~\cite{Szegedy_2016} is a network architecture that has been widely used, studied and outperforms a large number of competing architectures.
        A comparable baseline for segmenting remote sensing images does not exist.
\end{enumerate}
We consider \acrshort{fid} still as an experimental metric for evaluating performance and further research is needed.

\begin{table*}[htb]
    \caption{\Acrlong{miou}, pixel accuracies, and \acrshort{fid} when synthesizing RGB images, either from a single input or by fusing inputs using concatenation at the input, the merging generator or the proposed method.
    Fusing \acrshort{dem} (GeoNRW) or \acrshort{sar} (DFC2020) with land cover maps improves performance.
    We also compare against the segmentation maps obtained from real data to account for label noise.
    Label noise is also the reason why synthesizing from labels alone is advantageous when comparing to the ground truth.
}%
\label{tab_miou_pix_acc_fid}
\centering
\newcommand{\ra}[1]{\renewcommand{\arraystretch}{#1}}
\newcommand{\rotentry}[2]{\parbox[t]{2mm}{\multirow{#1}{*}{\rotatebox[origin=c]{90}{#2}}}}
\ra{1.2}
\begin{tabular*}{\textwidth}{@{\extracolsep{\fill}}llrrrrrrrrrrr}
\toprule
Dataset &            & Real     & \multicolumn{5}{c}{Fake with respect to Ground Truth}           & \multicolumn{5}{c}{Fake with respect to Real} \\
                                  \cmidrule{4-8}                                                    \cmidrule{9-13}
        &            &          &  DEM/SAR   &  Label      & Concat & Merge & Ours               &  DEM/SAR & Label & Concat & Merge & Ours \\
\midrule
GeoNRW  & mIoU       & 0.1734   & 0.1845          & \textbf{0.1991} & 0.1706 & 0.1706 & 0.1836             & 0.2315        & 0.1999     & 0.2317 & 0.2201 & \textbf{0.2326} \\
        & pixel acc. & 0.5434   & 0.5509          & 0.5367          & 0.5470 & 0.5470 & \textbf{0.5617}    & 0.7528        & 0.6641     & 0.7638 & 0.7575 & \textbf{0.7692} \\
        & FID        & ---      & ---             & ---             & ---    & ---    & ---                & 0.0307        & 0.0233     & 0.0097 & 0.0223 & \textbf{0.0078} \\

\midrule             
DFC2020 & mIoU       & 0.3722   & 0.3385          & \textbf{0.3802} & 0.3786 & 0.3787 & 0.3758             & 0.3846        & 0.3946     & 0.4352 & \textbf{0.4431} & 0.4344 \\
        & pixel acc. & 0.7514   & 0.7175          & \textbf{0.7661} & 0.7571 & 0.7602 & 0.7524             & 0.7778        & 0.7897     & 0.8209 & \textbf{0.8243} & 0.8095 \\
        & FID        & ---      & ---             & ---             & ---    & ---    & ---              & \textbf{0.0035} & 0.0916     & 0.0643          & 0.0150 & 0.0174 \\
\bottomrule%
\end{tabular*}%
\end{table*}

\begin{figure}
    \centering
    \includegraphics[width=\columnwidth]{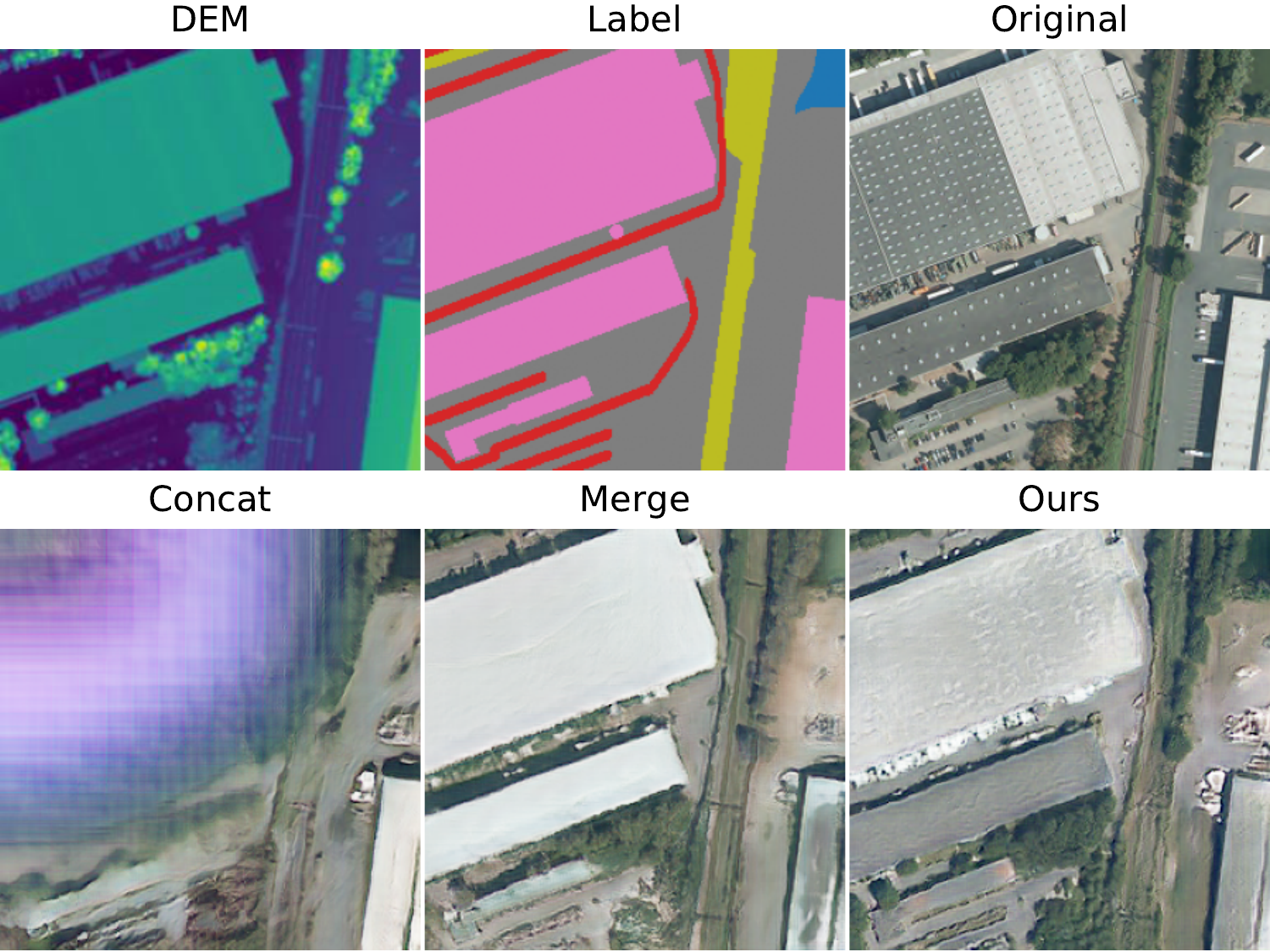}
    \caption{Comparison of the three different generator architectures for fusing \acrshort{dem} and land-cover maps. Simple concatenation in conjunction with a traditional produces artifacts for large buildings. The merging approach produces images with fewer details and washed-out textures due to its lower capacity.
}%
    \label{fig_nrw_concat_merge_ours}
\end{figure}

Looking at the numbers for the GeoNRW dataset in \Cref{tab_miou_pix_acc_fid}, simply concatenating inputs and using a traditional generator performs largely comparably to our method and both outperform the merging of essentially two generators by a third one.
\Cref{fig_nrw_concat_merge_ours} illustrates with an example some of the issues with the other approaches.
Concatenating and a traditional generator was prone to produce artifacts for large buildings.
Presumably the different magnitude of building heights and the one-hot encoded land-cover maps poses difficulties for their normalization.
Such distortions in only few samples of the dataset are not accurately reflected by global error metrics.
The merging approach with less network capacity produces outputs with fewer details and more washed-out textures, consistent with the findings in~\cite{brock_biggan_2019}.

The results for the DFC2020 dataset are more ambiguous, where the merging generator outperforms the other approaches.
One reason might be, that its lower capacity, which leads to outputs with fewer details, is not as relevant for medium-resolution as for high-resolution data.
We also conjecture that the pre-trained segmentation networks play a major role and might be more sensitive to missing details for the high-resolution GeoNRW dataset.
As mentioned before, subsequent research needs to find network architectures and training procedures so that these metrics are as reliable as in computer vision.

\subsection{Editing the input}

One of the motivations of this research is to provide a glimpse into the future.
For example, flooding or rising sea levels can be simulated, or the conversion of forests into farmland.

\begin{figure*}
    \centering
    \includegraphics[width=\textwidth]{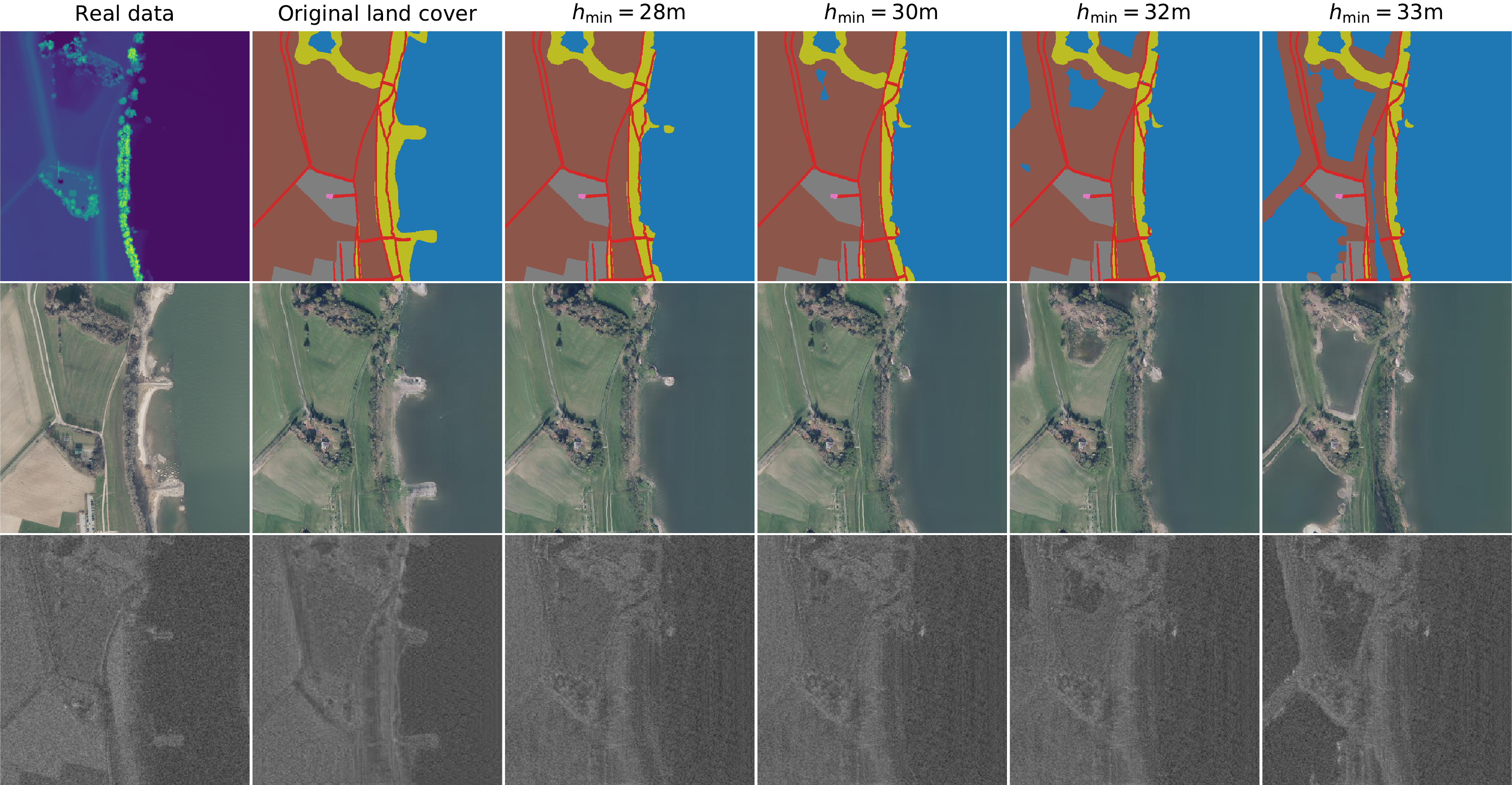}
    \caption{We threshold the \gls{dem} and adjust the land cover map to simulate flooding or rising water levels.
    The first column shows the original RGB, \acrshort{sar}, and \acrshort{dem} data.
The other columns show land cover maps and the corresponding synthesized images, where the first result corresponds to the original land cover map.
Light editing leads to realistically looking results.
For bigger changes the corresponding \gls{dem} and land cover map, computed by simple thresholding, probably do not adhere to the real data distribution anymore, resulting in slightly odd looking results.
}%
    \label{fig_nrw_flooding}
\end{figure*}

\Cref{fig_nrw_flooding} shows synthesized RGB and \acrshort{sar} images, for slightly altered inputs to the generator, trained on the GeoNRW dataset.
Thresholding the \acrshort{dem} with a minimum height $h_{min}$ creates a binary mask, which is subsequently cleaned up using a morphological erosion to get rid of small perturbations.
We assign the class water to all masked pixels and set the corresponding heights of the \acrshort{dem} to the thresholding value.
We feed the such created land cover maps and digital elevation models as inputs to the generator, simulating rising water levels.

For both RGB and \acrshort{sar} changing the segmentation map and \acrshort{dem} results in expanding water bodies.
The RGB image exhibits even realistic details like sand at the shoreline.

\section{Conclusion and Future Work}

With the proposed generator architecture and datasets, synthesizing convincing remote sensing imagery from abstract land cover maps and additional raster data, largely indistinguishable from real images, is possible.
What the work is currently lacking is more control on the results.
Generating the same scenery under different weather conditions, seasons or geographic locations, or additionally for \acrshort{sar} with different acquisition angles, are some of our future research directions.
These can be helpful for providing a variety of training data for other machine learning algorithms.
We are particularly interested in synthesizing images of disasters, where we need to take into account the domain gap between pre- and post-disaster images, as well as their imbalance.
For example, flooded areas look quite different from lakes and rivers, since the water is mixed with soil, or adding landslides and debris at certain locations.
Analysing the differences of \acrshort{sar} speckle statistics for real and synthesized data and see whether it matches theoretical considerations might also be a worthwhile topic.

\section*{Acknowledgment}
\addcontentsline{toc}{section}{Acknowledgment}

This work was supported in part by the Japan Society for the Promotion of Science through KAKENHI under Grants 18K18067 and 20K19834.
TerraSAR-X spotlight data were provided by DLR within in scope of the research proposal MTH3726: “SAR image synthesis from digital elevation models and land cover maps”.
Map data copyrighted OpenStreetMap contributors and available from \url{https://www.openstreetmap.org}.

\printbibliography%

\end{document}